\documentclass{jfpda}

\usepackage[pdftex]{pstricks}
\usepackage{verbatim}
\usepackage{graphicx}
\usepackage[algo2e,vlined,ruled]{algorithm2e}
\usepackage{amssymb}
\usepackage{amsmath}
\usepackage{epstopdf}
\usepackage{natbib}

\shorttitle{Q-Learning for Autonomous Soaring}

\shortouvrage{JFPDA 2017}

\title{Empirical evaluation of a Q-Learning Algorithm for Model-free Autonomous Soaring}

\author{Erwan Lecarpentier\inst{1}, Sebastian Rapp\inst{2}, Marc Melo, Emmanuel Rachelson\inst{3}}

\institute{
	ONERA -- DTIS (Traitement de l'Information et Syst\`{e}mes) \\
	2 avenue Edouard Belin, 31000 Toulouse, France \\
	\texttt{erwan.lecarpentier@isae.fr}
	\and
	TU Delft -- Department of Aerodynamics, Wind Energy \& Propulsion \\
	Building 62, room B62-5.07, Kluyverweg 1, 2629 HS Delft, Netherlands \\
	\texttt{s.rapp@tudelft.nl}
	\and
	ISAE Supaero -- DISC (D\'{e}partement d'Ing\'{e}nierie des Syst\`{e}mes Complexes)\\
	10 avenue Edouard Belin, 31055 Toulouse, France\\
	\texttt{emmanuel.rachelson@isae.fr}
}

\begin{document}

\maketitle

\begin{abstract}
Autonomous unpowered flight is a challenge for control and guidance systems: all the energy the aircraft might use during flight has to be harvested directly from the atmosphere. We investigate the design of an algorithm that optimizes the closed-loop control of a glider's bank and sideslip angles, while flying in the lower convective layer of the atmosphere in order to increase its mission endurance. Using a Reinforcement Learning approach, we demonstrate the possibility for real-time adaptation of the glider's behaviour to the time-varying and noisy conditions associated with thermal soaring flight. Our approach is online, data-based and model-free, hence avoids the pitfalls of aerological and aircraft modelling and allow us to deal with uncertainties and non-stationarity. Additionally, we put a particular emphasis on keeping low computational requirements in order to make on-board execution feasible. This article presents the stochastic, time-dependent aerological model used for simulation, together with a standard aircraft model. Then we introduce an adaptation of a $Q$-learning algorithm and demonstrate its ability to control the aircraft and improve its endurance by exploiting updrafts in non-stationary scenarios.

\end{abstract}

\section{INTRODUCTION}

The number of both civil and military applications of small unmanned aerial vehicles (UAVs) has augmented during the past few years. However, as the complexity of their tasks is increasing, extending the range and flight duration of UAVs becomes a key issue. Since the size, and thus the energy storage capacity, is a crucial limiting factor, other means to increase the flight duration have to be examined. A promising alternative is the use of atmospheric energy in the form of gusts and updrafts. This could significantly augment the mission duration while simultaneously save fuel or electrical energy. For this reason, there is a great interest in the development of algorithms that optimize the trajectories of soaring UAVs by harvesting the energy of the atmosphere. Since the atmospheric conditions are changing over time, it is crucial to develop an algorithm able to find an optimal compromise between exploring and exploiting convective thermal regions, while constantly adapting itself to the changing environment.

In this work we adapt a $Q$-learning \citep{watkins92qlearning} algorithm for this task. Our method is model-free, therefore suitable for a large range of environments and aircraft. Additionally, it does not need pre-optimization or pre-training, works in real-time, and can be applied online. Although the gap towards a fully autonomous physical demonstrator has not been bridged yet, our main contribution in this work is the \emph{proof of concept} that a model-free reinforcement learning approach can efficiently enhance a glider's endurance.
We start by reviewing the state of the art in UAV static soaring and thermal modelling in Section \ref{sec:relwork} and position our contributions within previous related work. Then, in Section \ref{sec:atmos}, we present the specific atmospheric model we used and its improvements over previous contributions, along with the thermals scenario used in later experiments. Section \ref{sec:aircraft} details the aircraft dynamics model. We introduce our implementation of the $Q$-learning algorithm in Section \ref{sec:control} and discuss its strengths, weaknesses and specific features. Simulation results are presented in Section \ref{sec:results}. We finally discuss the limitations of our approach and conclude in Section \ref{sec:conclu}.

\section{RELATED WORK}
\label{sec:relwork}

During the last decade, several possibilities to efficiently utilize atmospheric energy for soaring aircraft have been proposed. For a general introduction to static and dynamic soaring, refer to \citet{chen1981} for instance. For a more specific review on thermal centring and soaring in practice, see \citet{reichmann}.

Most approaches to thermal soaring rely on the identification of some model of the wind field surrounding the aircraft. This estimated wind field is then used to track an optimized trajectory inside the thermal or between thermals, using various methods for identification and path planning \citep{allen05,allen07,lawrance11,lawrance_phd,bencatel13,chen11,chakrabarty}. Such approaches demonstrated important energy savings (up to 90\% in simulation \citep{chakrabarty}) compared to conventional flight. An alternative robust control algorithm \citep{kahveci}, based again on a pre-identification of a thermal model showed good results also.

In this paper, we reconsider the possibility to use a \emph{Reinforcement Learning} \citep[RL, ][]{sutton_book} approach to optimize the trajectory. Using RL to exploit thermals has already been examined by \citet{wharington_phd}. In this work, a neural-based thermal centre locator for the optimal autonomous exploitation of the thermals is developed. After each completed circle, the algorithm memorizes the heading where the lift was the strongest and moves the circling trajectory towards the lift. However, this thermal locator is too time consuming for real-time on-board applications.

We introduce a \emph{Q-learning} algorithm using a \emph{linear function approximation}, which is simple to implement, demands less computational resources and does not rely on the identification of a thermal model.
We empirically evaluate this online learning algorithm (Section \ref{sec:control}) by interfacing it with a simulation model that couples the aircraft dynamics (Section \ref{sec:aircraft}) with an improved local aerological model (Section \ref{sec:atmos}).
We use the model to test our algorithm in several scenarios and show that it yields a significant endurance improvement. Our algorithm's main feature lies in its complete independence of the characteristics of the aerological environment, which makes it robust against model inaccuracy and estimation noise. Moreover, not explicitly estimating the thermal centre position and updraft magnitude saves valuable computational time.

\section{ATMOSPHERIC MODEL}
\label{sec:atmos}

Our updraft model expands on that of \citet{allen_thermal}. His model possesses three desirable features: dependence of the updraft distribution in the vertical direction, explicit modelling of downdrafts at the thermal's border and at every altitude, and finally the use of an environmental sink rate to ensure conservation of mass. Although a complete literature review on modelling the convective boundary layer is beyond the scope of this paper, it should be noted that \citet{allen_thermal} is the first reference that includes these three modelling aspects.


\begin{figure}
\begin{center}
 \includegraphics[width=9cm]{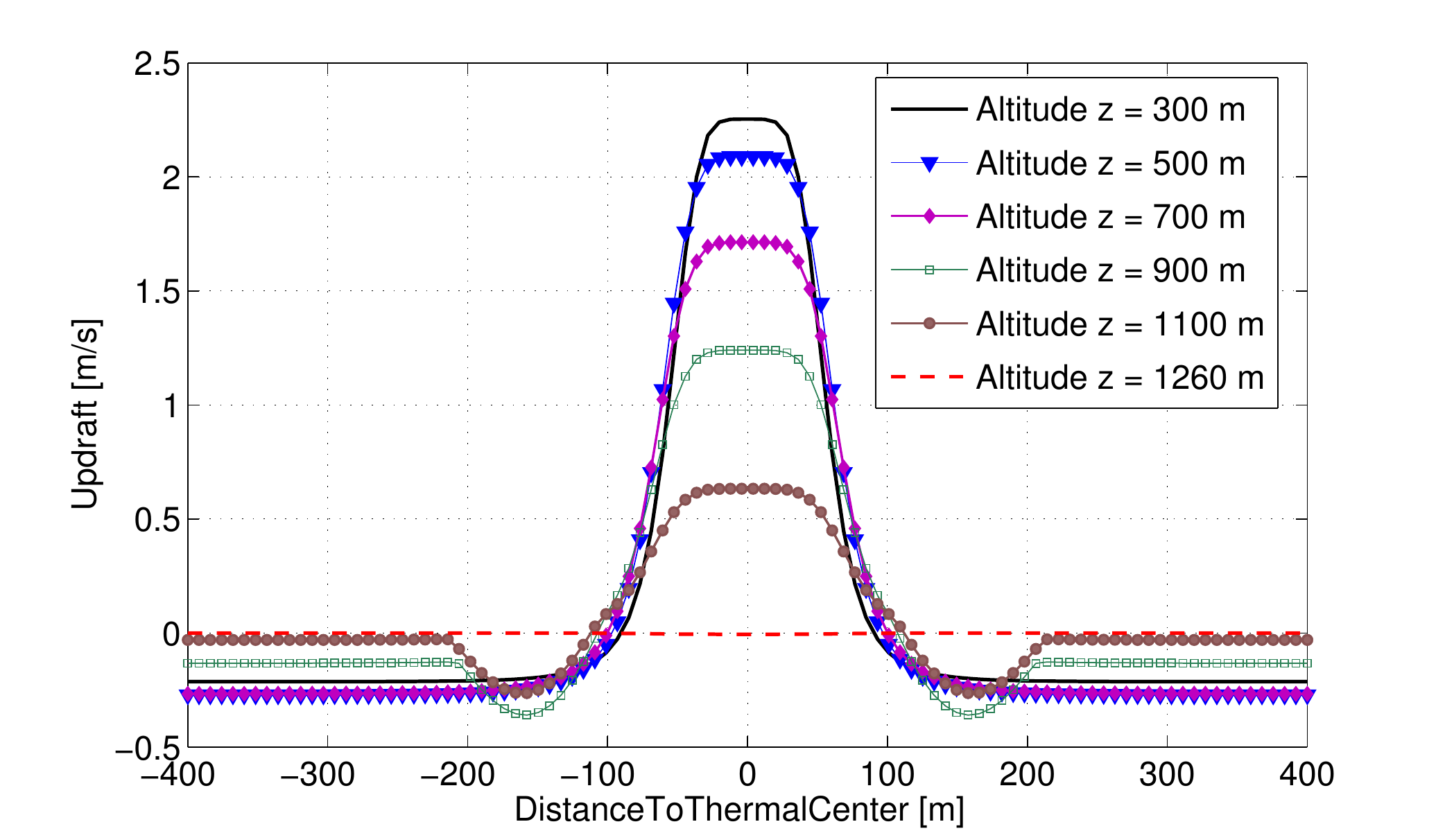}
\end{center}
 \caption{Updraft distribution with altitude}
 \label{fig:updraft_distribution}
\end{figure}


We describe a thermal updraft as a symmetrical, bell-shaped distribution as illustrated in Figure \ref{fig:updraft_distribution}. This distribution is characterized by two radii $r_1$ and $r_2$. At a given altitude $z$, if $r$ denotes the distance to the thermal center, for $r<r_1$ the updraft has a quasi-constant value of $w_{peak}$, then for $r_1<r<r_2$ this value drops smoothly to zero, and between $r_2$ and $2r_2$ appears a downdraft. The thermal has no influence further than $2r_2$.

The maximum updraft velocity $w_{peak}$ evolves altitude-wise proportionally to $w^* \left( \frac{z}{z_i} \right)^{\frac{1}{3}} \left(1 - 1.1 \ \frac{z}{z_i}\right)$, where $w^*$ is an average updraft velocity and $z_i$ is a scaling factor indicating the convective boundary layer thickness.
Above $0.9z_i$ all velocities are assumed to be zero.

Finally, based on the conservation of mass principle, an altitude-dependent global environmental sink rate is calculated and applied everywhere outside the thermals. For specific equations, we refer the reader to \citet{allen_thermal}.

\begin{figure}
\begin{center}
 \includegraphics[width=9cm]{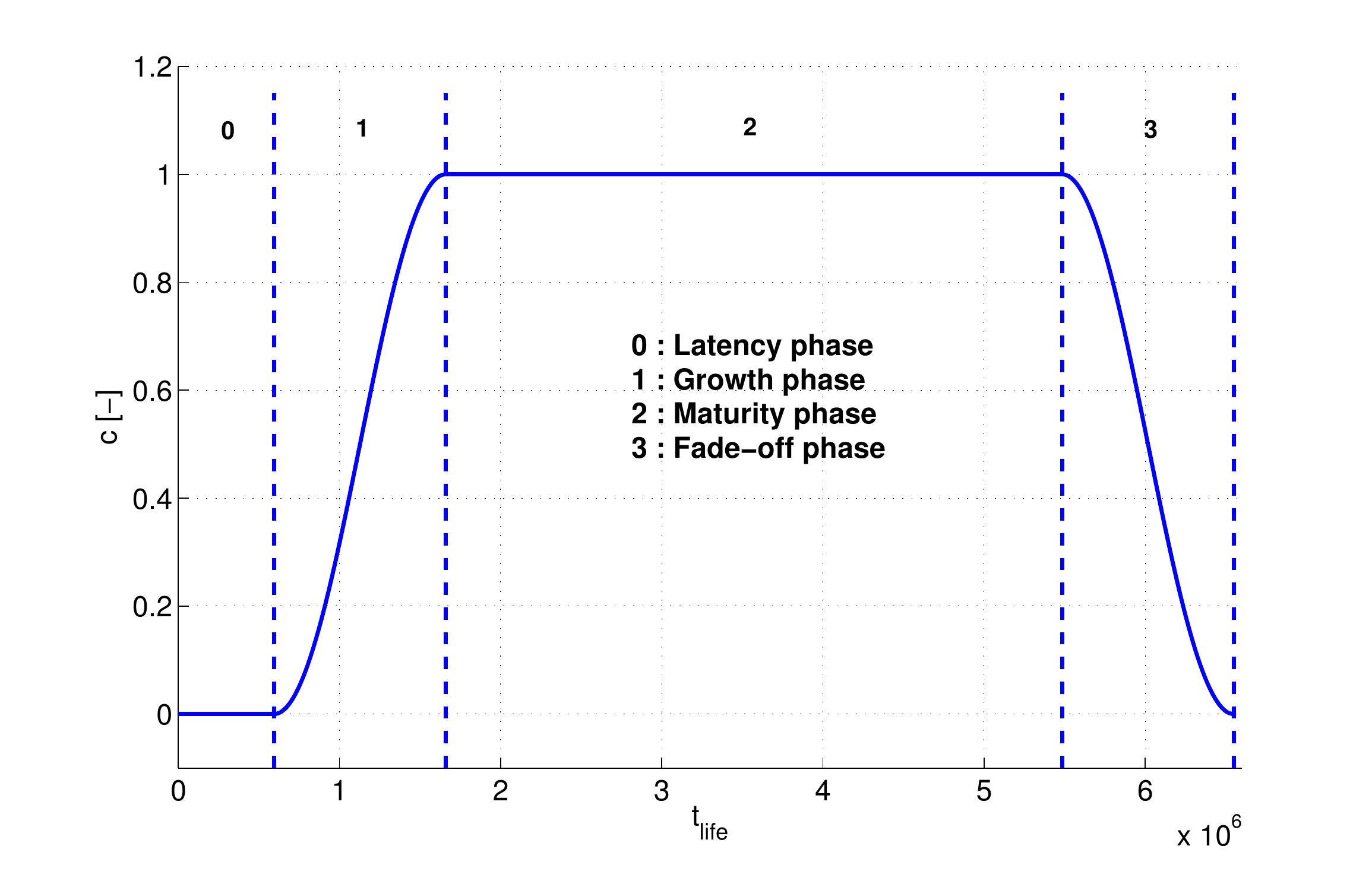}
\end{center}
\caption{Evolution of the updraft coefficient $c_\xi(t)$}
\label{fig:life_cycle}
\end{figure}

We introduce three additional features that bring our simulation model closer to a real-life description, namely thermal drift, life-cycle and noise.
First, in order to account for local winds, we let the thermals drift in the horizontal plane with a velocity $(\bar{v}_x, \bar{v}_z)$. Usually, the root point of a thermal is a fixed location and the thermal leans with the wind, so introducing a thermal drift is a poor description of this phenomenon. Nevertheless, for our simulations, it approximates the practical phenomenon of drift given that the aircraft model is reduced to a single point-mass.
Thermals also have a finite life. We decompose a thermal's life in a latency phase of duration $t_{\textit{off}}$ and a growth, maturity and fade-off phase of duration $t_{\textit{life}}$. After $t_{\textit{off}} \, + \, t_{\textit{life}}$ the thermal dies. The life-cycle of a thermal is described by the updraft coefficient $c_\xi(t)$ shown in Figure \ref{fig:life_cycle}, using a shape parameter $\xi$. This $c_\xi(t)$ coefficient is used as a multiplier on the total updraft.
Finally, it is well-known among cross-country pilots that thermals are rarely round and present a great variety of shapes and much noise. In order to account for this fact and to model real-life uncertainties we added a Gaussian distributed noise $n$ to the wind velocity.

We maintain a constant number $N$ of thermals in the flight area, although some might be in their latency phase. Consequently, whenever a thermal dies, a new thermal is generated with randomly drawn parameters $\{x_{th},y_{th}, w^*, z_i, \bar{v}_x, \bar{v}_y, t_{\textit{off}}, t_{\textit{life}}, \xi \}$.


\section{AIRCRAFT MODEL}
\label{sec:aircraft}

To model the dynamical behaviour of our aircraft, we used the equations derived by \citet{dynamic}, which consider the aircraft as a point-mass, 6 degrees of freedom system, and take into account the three dimensional wind velocity vector of the atmosphere as well as a parametric model for the aircraft's aerodynamics.
Let $m$ be the glider's mass and $g$ the gravity acceleration. The used variables are:
\begin{itemize}
	\item $x, y, z$ the coordinates in the earth frame;
	\item $V$ the absolute value of the aircraft's velocity in the earth frame;
	\item $\gamma$ the angle of climb;
	\item $\chi$ the course angle;
	\item $\alpha$ the angle of attack;
	\item $\beta$ the sideslip angle;
	\item $\mu$ the bank angle;
	\item $L, D $ and $C$ the lift, drag and lateral force.
\end{itemize}
The corresponding equations are described below:
\begin{align*}
\dot{x} &= V \cos(\chi)\cos(\gamma)\\
\dot{z} &= V \sin(\gamma)\\
\dot{y} &= V \sin(\chi)\cos(\gamma)\\
\dot{V} &= -\frac{D}{m}-g \sin(\gamma)\\
\dot{\gamma} &= \frac{1}{mV}\left(L\cos(\mu) + C \sin(\mu) - \frac{g}{V}\cos(\gamma)\right)\\
\dot{\chi} &= \frac{1}{mV \cos(\gamma)}\left(L\sin\left(\mu\right)-C \cos\left(\mu\right)\right)
\end{align*}
The first three equations describe the kinematics and position rates in the earth frame.
The last three equations define the dynamics of the glider aircraft.
For a detailed presentation of the aerodynamic parameters and forces, we refer the reader to \citet{dynamic}.
Adopting this modelling directly implies taking the three angles $\alpha, \, \beta$ and $\mu$ as control variables. Indeed the lift force depends on the bank angle, while the drag and lateral force depend on the three angles.
For simplicity of notations we omitted to write this dependency in the model's equations.
The choice of the state and action spaces considered by the controller is discussed in Section \ref{sec:control:stateaction}.

\section{ADAPTIVE CONTROLLER}
\label{sec:control}

\subsection{$Q$-learning}

RL \citep{sutton_book} is a branch of Discrete-time Stochastic Optimal Control that aims at designing optimal controllers for non-linear, noisy systems, using only interaction data and no \emph{a priori} model. The only hypothesis underlying RL algorithms is that the system to control can be modelled as a Markov Decision Process \citep[MDP, ][]{puterman}, even if this model is not available. An MDP is given by a set of system states $s\in S$, a set of control actions $a\in A$, a discrete-time transition function $p(s'|s,a)$ denoting the probability of reaching state $s'$ given that action $a$ was undertaken in state $s$, and finally a reward model $r(s,a,s')$ indicating how valuable the $(s,a,s')$ transition was with respect to the criterion one wants to maximize.

The overall goal of an RL algorithm is to derive an optimal control policy $\pi^*(s) = a$ that maximizes the expected cumulative sum of rewards $\mathbb{E}\left(\sum_{t=0}^\infty \eta^t r_t\right)$ from any starting state $s$ ($\eta\in[0;1[$ being a discount factor over future rewards). We focus on model-free RL algorithms that do not commit to the knowledge of the transition and reward models of the underlying MDP but use \emph{samples} of the form $(s,a,r,s')$ to learn an optimal policy.
In our case, that means that an RL algorithm controlling the glider with an overall goal of gaining energy will use sensor data to build $\pi^*$ online, without relying on a (possibly approximate) model of the atmosphere, or the aircraft's flight dynamics.

$Q$-learning, introduced by \citet{watkins92qlearning}, is one of the most simple and popular online RL algorithms. It aims at estimating the optimal action-value function $Q^*(s,a)$ in order to asymptotically act optimally. This function denotes the expected gain of applying action $a$ from state $s$, and then applying an optimal control policy $\pi^*$:
\begin{equation*}
Q^*(s,a) = \mathbb{E}\left(\sum\limits_{t=0}^\infty \eta^t r_t | s_0=s, a_0=a, a_t=\pi^*(s_t)\right)
\end{equation*}
The key idea behind $Q$-learning is that the optimal action in state $s$ is the one that maximizes $Q^*(s,a)$. Thus the optimal policy is greedy with respect to $Q^*$ in every state.
Estimating $Q^*$ from $\left(s,a,r,s'\right)$ samples is a stochastic approximation problem which can be solved with a procedure known as \emph{temporal differences}. The $Q$-learning algorithm is summarized in Algorithm \ref{alg:q-learning}.

\begin{algorithm2e}
\DontPrintSemicolon
Initialize $Q(s,a)$ for all $(s,a) \in S \times A$,\;{}
$s_{t} \leftarrow s_{0}$.\;
\Repeat{simulation end}{
	Apply $a_{t}=\arg\max_{a \in A} Q(s_t,a)$ with probability $1-\epsilon_t$, otherwise apply a random action $a_{t}$ \;
	Observe $s_{t+1}$ and $r_{t}$ \;
	$\delta_{t} = r_{t} + \eta \ \max_{a' \in A} \left(Q\left(s_{t+1},a'\right)\right) - Q\left(s_t, a_t\right)$ \;
	Update $Q(s_{t},a_{t}) \leftarrow Q(s_t,a_t) + \alpha_t \delta_t$ \label{eq:Qupdate} \;
	$s_{t}\leftarrow s_{t+1}$
}
\caption{$Q$-learning}
\label{alg:q-learning}
\end{algorithm2e}

Notice that $Q$-learning is an \emph{off-policy} method, that is, it estimates $Q^*$ assuming that a greedy policy w.r.t. $Q$ is followed.
However, the undertaken action at time $t$ is not necessarily greedy and can be randomly chosen with probability $\epsilon_t$.
This strategy, so-called \emph{$\epsilon$-greedy}, allows a wider exploration of the state-action space granting a better estimation of the $Q$-function.
As $\epsilon_t$ tends towards zero, if the learnt $Q$-function has converged to $Q^*$, the agent tends to act optimally.
As long as all state-action pairs are visited infinitely often when $t\rightarrow\infty$, $Q$ is guaranteed to converge to $Q^*$ if the sequence of learning rates $\alpha_t$ satisfies the conditions of \citet{robbins1951}: 
\begin{equation*}
\sum_{t=0}^\infty \alpha_t = \infty, \ \
\sum_{t=0}^\infty \alpha_t^2 < \infty
\end{equation*}
In the remainder of this section, we discuss how our problem differs from the vanilla MDP and $Q$-learning frameworks, and the design choices we made to accommodate these differences.

\subsection{State and action spaces}
\label{sec:control:stateaction}

Recall that the state of the aircraft, as defined in Section \ref{sec:aircraft}, or the state of the atmospheric model (Section \ref{sec:atmos}) are not fully observable to our learning agent. So it would be unrealistic to define the state space $S$ as the observations of these values. Instead, we suppose that a state only defined by $(\dot{z}, \dot{\gamma}, \mu, \beta)$ is accessible and that its dynamics still define an MDP. Such a state is easily measurable with reliable sensors such as pressure sensors, accelerometers or gyrometers. This key assumption is crucial to the success of our method since it reduces the size of the state space, easing the approximation of $Q^*(s,a)$. We shall see later that this choice of state variables has other advantages.

The considered actions consist in directly controlling the aircraft's bank and sideslip angles increments so that the action space is $A = \left\{-\delta\mu,0,\delta\mu\right\}\times\left\{-\delta\beta,0,\delta\beta\right\}$, resulting in $|A| = 9$ different possible actions.
We chose the values of $\delta\mu$ and $\delta\beta$ so that, given a certain control frequency, the cumulated effect of a constant action does not exceed the admissible dynamics of the aircraft. This results in a steady state change, representative of the actual behaviour of the actuators.

\subsection{Reward model}

The goal of our learning algorithm is to maximize the glider's endurance. This boils down to maximizing the expected total energy gain, so we wish that $Q(s,a)=\mathbb{E} \{ \textrm{total energy at }t=\infty \} $. To achieve this, we choose:
\begin{equation}
r_{t} = \dot{E}_{aircraft} = \frac{d}{dt} \left( z + \frac{V^2}{2g}\right)
\end{equation}
Thus we assume that this reward signal $r_t$ is provided to the learning algorithm at each time step, representing the (possibly noisy) total energy rate of the aircraft. Note that the variables $\dot{z}$, $V$ and $\dot{V}$ can be measured online with classical sensors such as a GPS and an accelerometer.

\subsection{Convergence in unsteady environments}

The previous requirements on $\epsilon_t$ and $\alpha_t$ for convergence of $Q$ to $Q^*$ hold if the environment can indeed be modeled as an MDP.
However, in the studied case, the environment is non-stationary since the thermals have a time-varying magnitude (thermal coefficient) and location (drift). Moreover, given the choice of state variables, since the agent is blind to its localization, the distribution $p(s'|s,a)$ is not stationary and changes from a time step to the other.
Consequently, our learning agent evolves in a constantly changing environment which is \emph{not} a stationary MDP and we actually need to rely on its ability to learn and adapt quickly to changing conditions if we wish to approximate these conditions as quasi-stationary. In order to allow this quick adaptation, we need to force a permanent exploration of the state-action space and to constantly question the reliability of $Q$. This corresponds to making use of constant $\alpha_t$ and $\epsilon_t$ values, which need to be well-chosen in order to retain a close-to-optimal behaviour while quickly adapting to the changes in the environment.

The choice of a simplified low-dimensional state space makes the adaptation to a non-stationary environment feasible. In fact, with our specific choice of state variables, in the short term, the learning agent observes a quasi-constant state $(\dot{z},\dot{\gamma}, \mu,\beta)$ and the optimal action in this state is almost constant as well. Indeed, the chosen variables evolve slowly through the time, making the evolution of the optimal action value slow as well. This allows to make maximal use	of the collected samples since only a local approximation around the current state is required to compute the current optimal action. The success of the method is therefore due to the capacity of the $Q$-learning algorithm to track the optimal action quickly enough in comparison to the environment's dynamics.

\subsection{Linear $Q$-function approximation}

In order to avoid the discretisation of the state space in the description of $Q$, we adopt a linear function approximation of $Q(s,a)$. We introduce sigmoid-normalized versions of the state space variables and define our basis functions $\phi$ as the monomials of these normalized variables of order zero to two (15 basis functions). Then, by writing $Q(s,a)=\theta^T \phi(s,a)$, the update equation of $Q$-learning becomes $\theta_{t+1}=\theta_t + \alpha_t\delta_t\phi(s_t,a_t)$. 
There is abundant literature on choice of feature functions in RL, we refer the reader to \citet{parr08}, \citet{hachiya10}, or \citet{nguyen13} for more details.

To summarize, our glider is controlled by a $Q$-learning algorithm with fixed learning and exploration rates ($\alpha$ and $\epsilon$) to account for the unsteadiness of the environment. The optimal action-value function $Q^{*}$ is approximated with a linear architecture of quadratic features defined over a set of observation variables $\left(\dot{z}, \dot{\gamma}, \mu, \beta \right)$. Finally, at each time step, the chosen action is picked among a set of 9 possible increments on the $\left(\mu, \beta\right)$ current values.

\section{SIMULATION RESULTS}
\label{sec:results}

We identify three scenarios designed to empirically evaluate the convergence rate of the algorithm and the overall behaviour of the glider. These scenarios take place within a 1100m wide circular flight arena. Whenever the glider exits the arena, an autopilot steers it back in. The aircraft is initialized at $z=300$m and $V=15$m/s. According to \citet{allen_thermal}, we set $w^*=2.56$m/s and $z_i=1401$m. The algorithm parameters were $\epsilon = 0.01$; $\alpha = 0.001$; $\eta = 0.99$; $\delta \beta = 0.003\deg$; $\delta \mu = 0.003\deg$; $\beta_{max} = 45\deg$; $\mu_{max} = 25\deg$ and the observation frequency is $1kHz$. 

\begin{figure}
\begin{center}
\includegraphics[width=9cm]{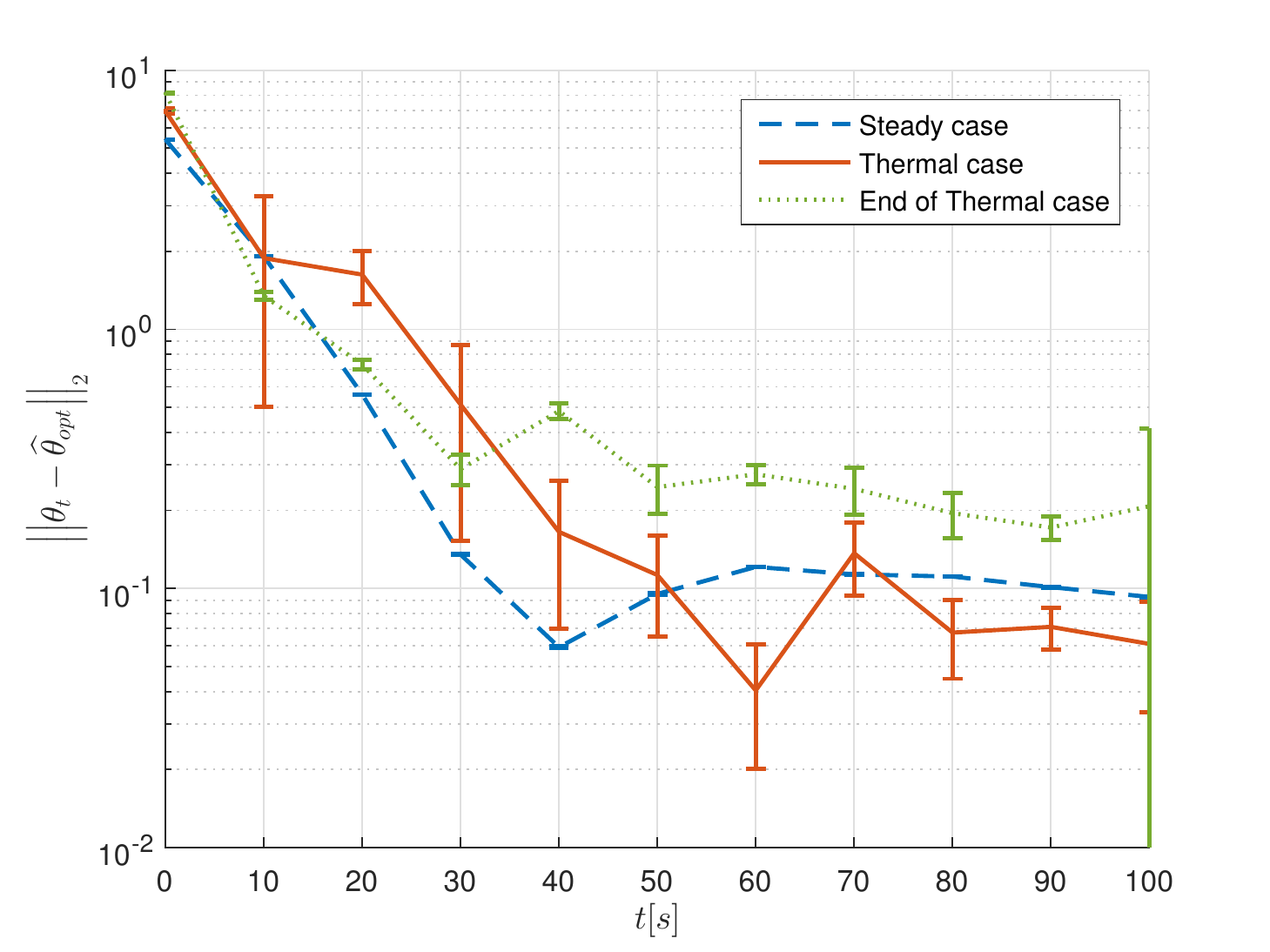}
\end{center}
\caption{Convergence of the action-value function}
\label{fig:param_cv}
\end{figure}

The three scenarios are the following: flight in still air without thermal but a noisy downdraft; birth of a thermal along the trajectory; death of a thermal into which the UAV was flying.
Qualitatively, the optimal policy in each case is respectively to adopt a straight flight configuration; to circle up within the thermal; and to switch from the circular trajectory to a straight one as in the first case.
In each scenario, we refer to the optimal action-value function parameters as $\theta_{opt}$.
In order to analyse the convergence rate of the algorithm, we built an empirical estimate $\widehat{\theta}_{opt}$ of those parameters with the value they take after the convergence of the algorithm and then compute the quantity $\|\theta_t - \widehat{\theta}_{opt}\|_2$ along 50 roll-outs of the system.
The convergence results are reported in Figure \ref{fig:param_cv} where the error bars indicate the standard deviation. One can see that the time required to adjust the parameters to each situation ranges between 30 and 40 seconds, which is compatible with the change rates of the glider's environment.
Note in particular that the glider's behaviour might be optimal long before $\theta$ converges to $\theta_{opt}$ since from a certain state $s$ the optimal action might be selected even if the parameters did not converge. Indeed, what matters is the ranking of the $Q$-values of the different actions rather than the $Q$-values themselves.
Practically, the configurations vary between the three studied cases and the exploratory feature of the $\epsilon$-greedy policy allows to permanently adapt the $Q$-function to the situation.

The performance reached by the control algorithm can be measured via the total energy of the aircraft, capturing the reached altitude and the velocity. In the three aforementioned scenarios, the expected results are not the same. Indeed, in a steady atmosphere, the optimal policy only allows to minimize the loss of altitude by setting $\beta = \mu = 0$. Such a configuration is optimal since no thermals can be found and the glider can only maximize its long term energy by flying straight and avoiding sharp manoeuvres. Then, when a thermal is reached, the algorithm's exploratory behaviour allows to captures the information that it is worth changing $\beta$ and $\mu$, and adapts the trajectory to maximize the long-term return. In the third situation, when the glider flies inside a dying thermal, the algorithm brings back the parameters to a steady atmosphere configuration and again minimizes the expected loss of energy.


\begin{figure}
\begin{center}
 \includegraphics[width=9cm]{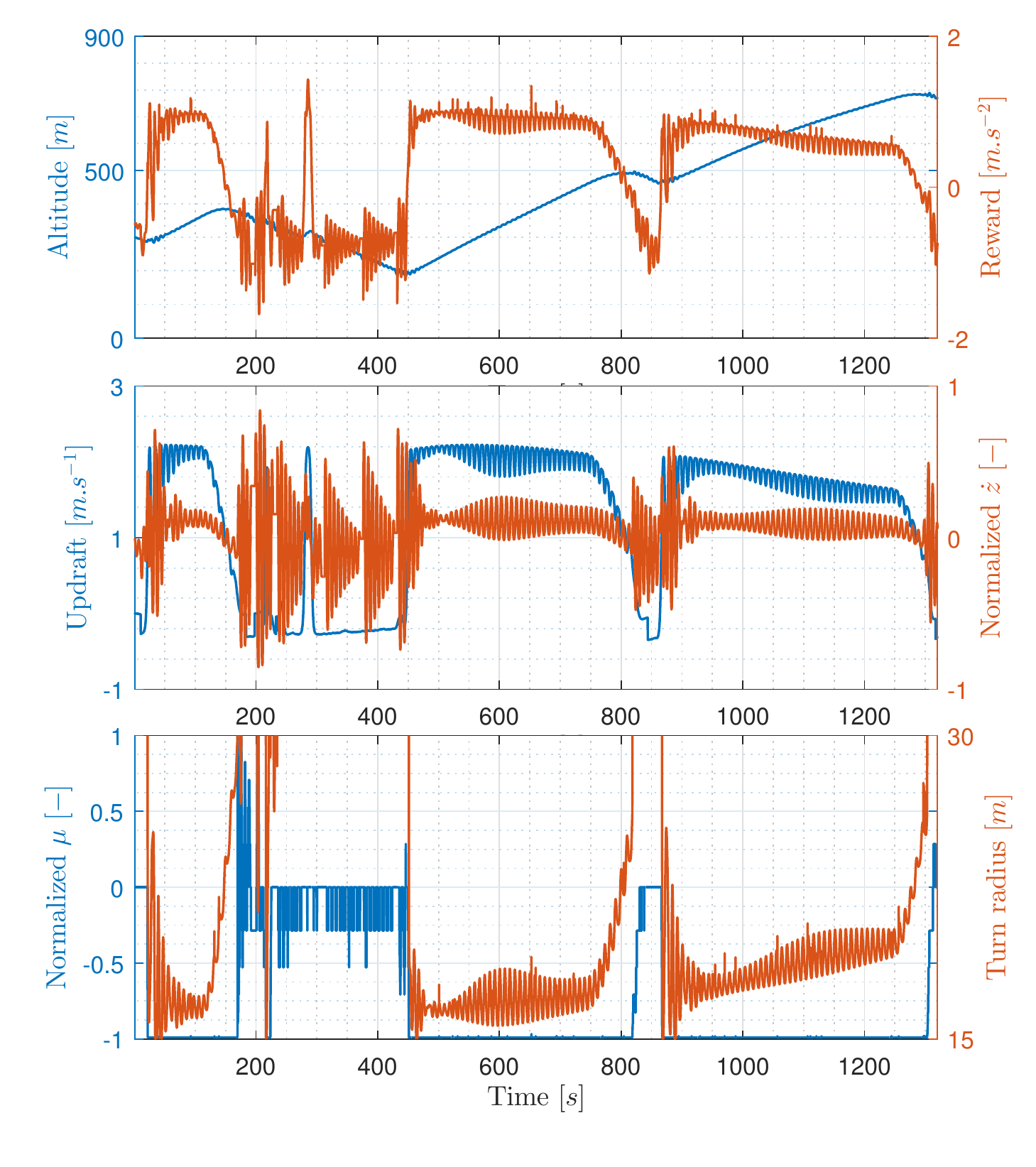}
\end{center}
\caption{Evolution of the aircraft variables with time}
\label{fig:traj_rho}
\end{figure}

\begin{figure}
\begin{center}
 \includegraphics[width=9cm]{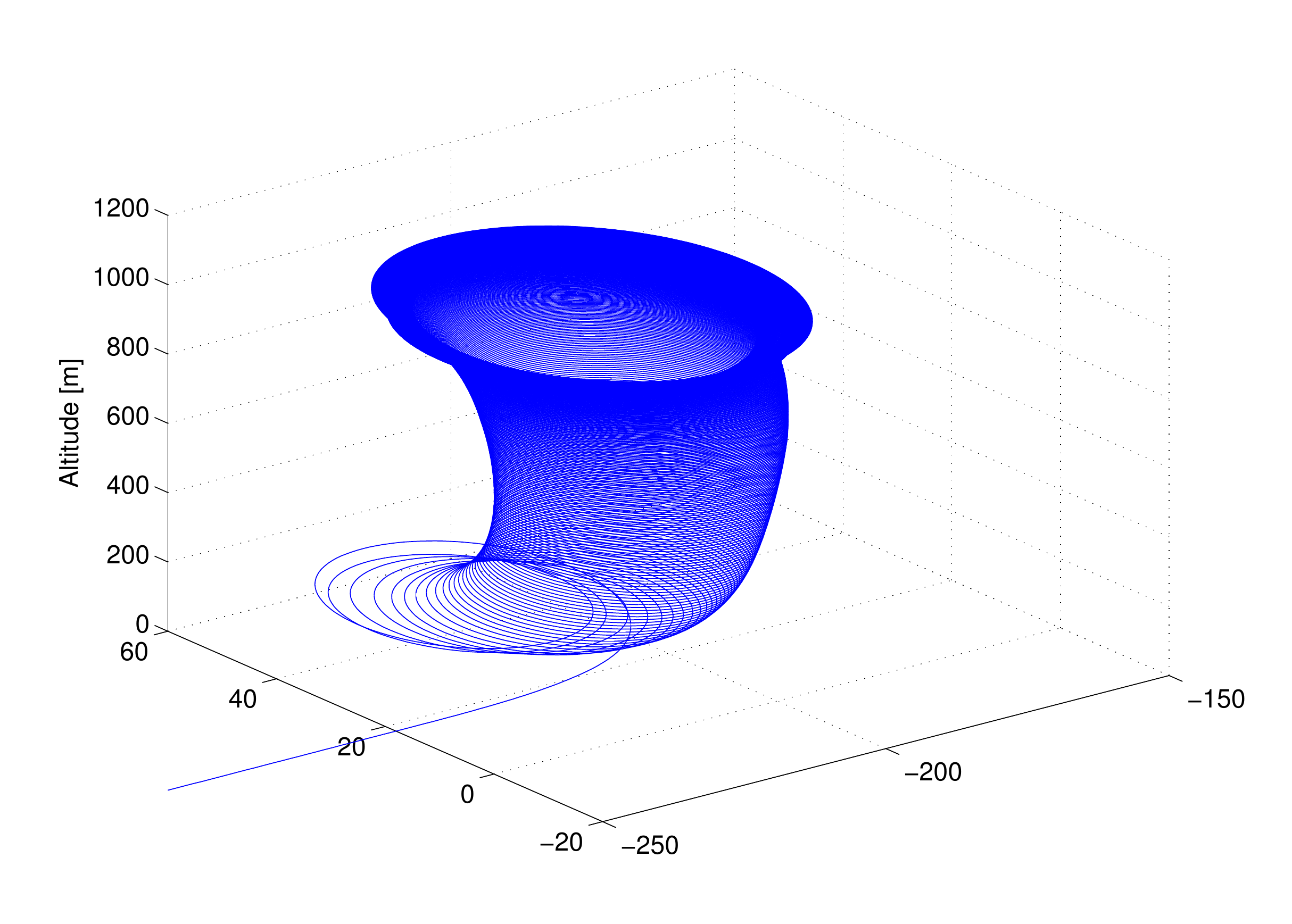}
\end{center}
 \caption{Trajectory of the aircraft inside a thermal}
 \label{fig:traj_high_alt}
\end{figure}

Figure \ref{fig:traj_rho} shows the evolution of altitude and instantaneous rewards through time in a typical long-term scenario with multiple thermal crossings. Each altitude pike shows the entry of the aircraft into a thermal. First the trajectory is bent in order to maximize the altitude gain and when the thermal dies, the glider goes back to the steady flight configuration. Clearly, each gain-of-altitude phase corresponds to a positive reward and, conversely, a loss-of-altitude phase to a negative one. A 3D display of the trajectory inside a thermal is presented in Figure \ref{fig:traj_high_alt}.

\begin{figure}
\begin{center}
 \includegraphics[width=9cm]{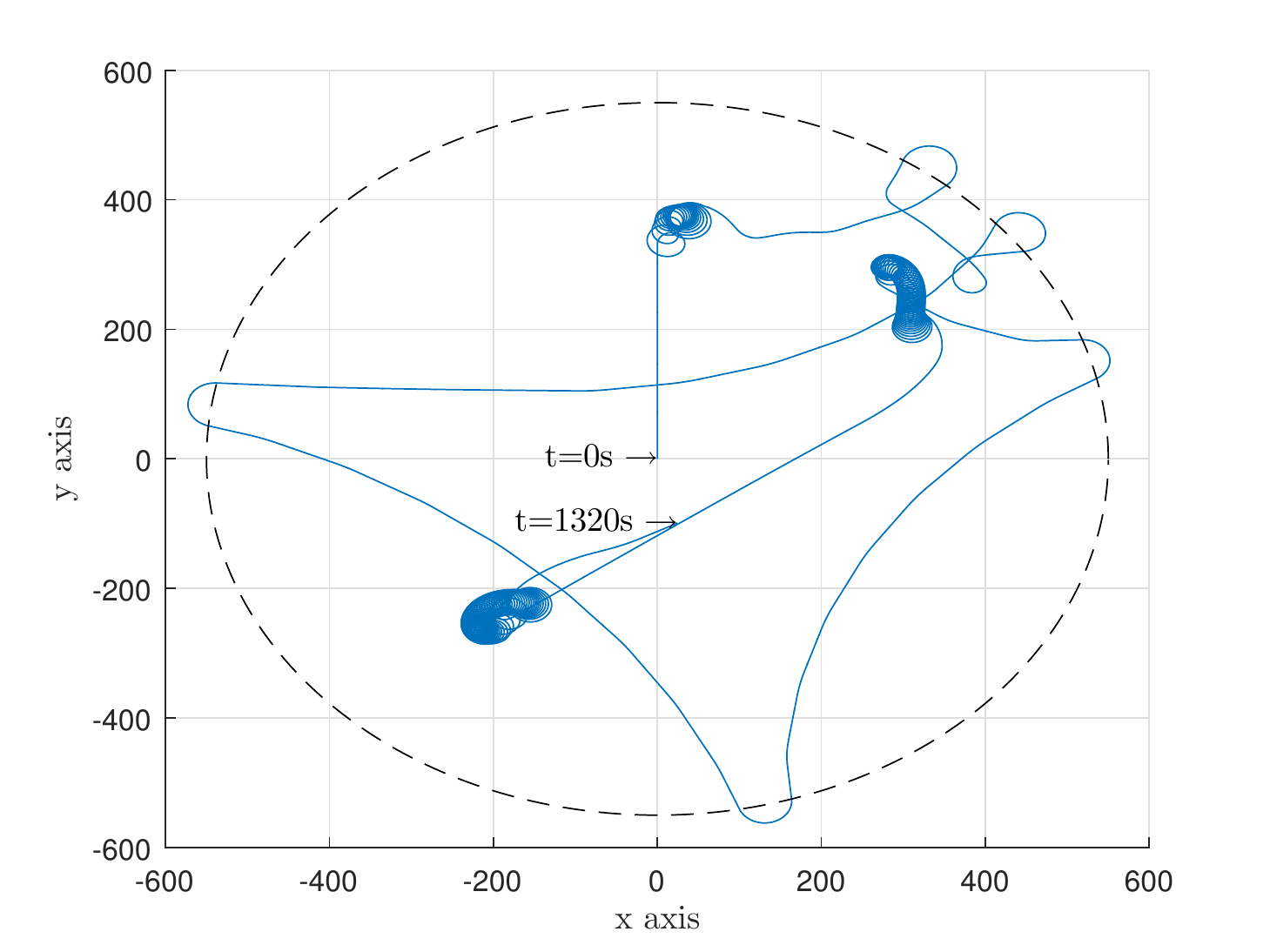}
\end{center}
\caption{An example of trajectory}
\label{fig:full_traj}
\end{figure}

The $Q$-learning controller yields an overall behaviour close to the one of a human pilot while being totally unaware of its own location and of local wind field models. When flying in still air, the glider remains in ``flat'' flight attitude, thus maximizing its flight time expectancy. Whenever an updraft is spotted, it engages in a spiral, as shown in Figure \ref{fig:traj_rho}. If the updraft dies, the aircraft comes back to the first configuration. This results in an overall trajectory composed with straight lines and circles as displayed in Figure \ref{fig:full_traj}.

Figure \ref{fig:traj_rho} also illustrates the reaction times of the glider and the overall command behaviour. It appears that the glider starts to circle up the thermals long before the value function has converged. Similarly, the convergence to a steady air optimal behaviour is faster than the $Q$-function convergence illustrated on Figure \ref{fig:param_cv}.
When the glider reaches the thermal's top, the updraft naturally decreases. Consequently one can notice the reduction of the bank angle (enlargement of the turning radius) computed by the algorithm in order to stay in the thermal while reaching a zero vertical velocity.


\section{DISCUSSION AND CONCLUSION}
\label{sec:conclu}

In this paragraph, we discuss the limitations of our contribution, highlight directions for improvements and underline how our results make a difference compared to related work in the literature presented in Section \ref{sec:relwork}. To summarize, we implemented a proof of concept that a computationally light algorithm like $Q$-learning could be adapted to take into account the time-varying conditions of thermal soaring flight and could make efficient online changes to the control behaviour of an autonomous glider. We take a critical look at this contribution.

First of all, we did not introduce a new RL algorithm \emph{per se}, even though we shortly discuss the question of learning in unsteady environments. The choice of $Q$-learning is justified by its low computational footprint, despite the existence of a vast literature of efficient algorithms in online RL. Our contributions on the RL side are application-specific: first we justify the need for constant $\alpha$ and $\epsilon$ parameters to account for permanent exploration and adaptation in our unsteady environments. Secondly, we make a particular choice of state and action variables, such that, under an optimal policy, the system remains in a quasi-constant state (it would not be the case if the coordinates $x,y,z$ were part of the state space for instance), thus limiting the need for exploration and making the learning process faster. Finally, we introduced a reward model based explicitly on the maximization of the long term energy of the aircraft, thus linking energetic considerations with the definition of the $Q$-function.

From a low-level control point of view, the hypothesis of a control frequency of 1kHz is somehow questionable and it should be decreased in further developments. We argue however that this frequency is representative of a measurement frequency and should thus still be used to update the $Q$-function. Exploratory actions artificially account for the information collected due to the noise in wind conditions felt by the aircraft.

The 6 degrees of freedom aircraft model used in the simulation is a classical flight dynamics model that does not take into account the wind gradient in the wingspan direction. This gradient however is known to be a crucial information for human pilots, since it disambiguates whether a thermal centre is on the left or right hand side of the glider. Exploiting such information could bring more efficiency to the glider's control and avoid missing some thermals because the turn was initiated in the wrong direction.

Lastly, in this proof of concept, we based the action space on the aerodynamic angles $\mu$ and $\beta$ as it was done by \citet{dynamic}. Since the $Q$-learning algorithm aims at maximizing the average energy gain in the long term, it does not improve the short-term stabilization of the longitudinal modes of the aircraft, leading to the oscillations shown in Figure \ref{fig:traj_rho}. Even though this does not affect the overall long-term energy gains, a desirable improvement would consist in implementing a low-level stabilization loop (with a PID controller for instance), thus allowing to define the action space using aircraft attitude set points, rather than aerodynamic angles.

Overall, our contribution is three-fold. First we report on how to efficiently adapt a $Q$-learning algorithm to the non-steady, partially observable, control problem of thermal soaring. Then we empirically evaluate the performance of this algorithm in a rich simulation environment, illustrating how it can be used to improve the energy autonomy of soaring planes. Finally we discuss the strengths and limitations of this approach, thus opening research perspectives on this topic and providing first insights on these perspectives.

\bibliography{mybiblio}

\begin{thebibliography}{~~~}

\bibitem[\protect\citename{Allen, }2005]{allen05}
{\sc Allen M.~J.} (2005).
\newblock {\em Autonmous Soaring for Improved Endurance of a Small Uninhabited
  Air Vehicle}.
\newblock Rapport interne, NASA Dryden Research Center.

\bibitem[\protect\citename{Allen, }2006]{allen_thermal}
{\sc Allen M.~J.} (2006).
\newblock {\em Updraft Model for development of Autonomous Soaring Uninhabited
  Air Vehicles}.
\newblock Rapport interne, NASA Dryden Flight Research Center.

\bibitem[\protect\citename{Allen \& Lin, }2007]{allen07}
{\sc Allen M.~J. \& Lin V.} (2007).
\newblock {\em Guidance and Control of an Autonomous Soaring UAV}.
\newblock Rapport interne, NASA Dryden Flight Research Center.

\bibitem[\protect\citename{Beeler {\em et~al.}, }2003]{dynamic}
{\sc Beeler S., Moerder D. \& Cox D.} (2003).
\newblock {\em A Flight Dynamics Model for a Small Glider in Ambient Winds}.
\newblock Rapport interne, NASA.

\bibitem[\protect\citename{Bencatel {\em et~al.}, }2013]{bencatel13}
{\sc Bencatel R., de~Sousa J.~T. \& Girard A.} (2013).
\newblock Atmospheric flow field models applicable for aircraft endurance
  extension.
\newblock {\em Prog. in Aerospace Sciences}, {\bf 61}.

\bibitem[\protect\citename{Chakrabarty \& Langelaan, }2010]{chakrabarty}
{\sc Chakrabarty A. \& Langelaan J.} (2010).
\newblock Flight path planning for {UAV} atmospheric energy harvesting using
  heuristic search.
\newblock In {\em AIAA Guidance, Navigation, and Control Conference}.

\bibitem[\protect\citename{Chen \& McMasters, }1981]{chen1981}
{\sc Chen M. \& McMasters J.} (1981).
\newblock From paleoaeronautics to altostratus - a technical history of
  soaring.
\newblock In {\em AIAA Aircraft Systems and Technology Conference}.

\bibitem[\protect\citename{Chen \& Clarke, }2011]{chen11}
{\sc Chen W. \& Clarke J. H.~A.} (2011).
\newblock Trajectory generation for autonomous soaring {UAS}.
\newblock In {\em 17th International Conference on Automation and Computing}.

\bibitem[\protect\citename{Hachiya \& Sugiyama, }2010]{hachiya10}
{\sc Hachiya H. \& Sugiyama M.} (2010).
\newblock Feature selection for reinforcement learning: Evaluating implicit
  state-reward dependency via conditional mutual information.
\newblock In {\em European Conference on Machine Learning and Knowledge
  Discovery in Databases}, p.\ 474--489.

\bibitem[\protect\citename{Kahveci \& Mirmirani, }2008]{kahveci}
{\sc Kahveci N. \& Mirmirani M.} (2008).
\newblock Adaptive {LQ} control with anti-windup augmentation to optimize {UAV}
  performance in autonomous soaring application.
\newblock In {\em IEEE Transactions on Control System Technology}.

\bibitem[\protect\citename{Lawrance, }2011]{lawrance_phd}
{\sc Lawrance N.} (2011).
\newblock {\em Autonomous Soaring Flight for Unmanned Aerial Vehicle}.
\newblock PhD thesis, The University Of Sydney.

\bibitem[\protect\citename{Lawrance \& Sukkarieh, }2011]{lawrance11}
{\sc Lawrance N. \& Sukkarieh S.} (2011).
\newblock Path planning for autonomous soaring flight in dynamic wind.
\newblock In {\em IEEE International Conference on Robotics and Automation}.

\bibitem[\protect\citename{Nguyen {\em et~al.}, }2013]{nguyen13}
{\sc Nguyen T., Li Z., Silander T. \& Leong T.~Y.} (2013).
\newblock Online feature selection for model-based reinforcement learning.
\newblock In {\em Int. Conf. on Machine Learning}.

\bibitem[\protect\citename{Parr {\em et~al.}, }2008]{parr08}
{\sc Parr R., Li L., Taylor G., Painter-Wakefield C. \& Littman M.~L.} (2008).
\newblock An analysis of linear models, linear value-function approximation,
  and feature selection for reinforcement learning.
\newblock In {\em International Conference on Machine Learning}.

\bibitem[\protect\citename{Puterman, }2005]{puterman}
{\sc Puterman M.~L.} (2005).
\newblock {\em Markov Decision Processes: Discrete Stochastic Dynamic
  Programming}.
\newblock John Wiley \& Sons, Inc.

\bibitem[\protect\citename{Reichmann, }1993]{reichmann}
{\sc Reichmann H.} (1993).
\newblock {\em Cross-Country Soaring}.
\newblock Soaring Society of America.

\bibitem[\protect\citename{Robbins \& Monro, }1951]{robbins1951}
{\sc Robbins H. \& Monro S.} (1951).
\newblock A stochastic approximation method.
\newblock {\em Ann. Math. Statist.}, {\bf 22}(3), 400--407.

\bibitem[\protect\citename{Sutton \& Barto, }1998]{sutton_book}
{\sc Sutton R.~S. \& Barto A.~G.} (1998).
\newblock {\em Reinforcement Learning: An Introduction}.
\newblock MIT Press.

\bibitem[\protect\citename{Watkins \& Dayan, }1992]{watkins92qlearning}
{\sc Watkins C. J.~C. \& Dayan P.} (1992).
\newblock {Q}-learning.
\newblock {\em Machine Learning}, {\bf 8}, 279--292.

\bibitem[\protect\citename{Wharington, }1998]{wharington_phd}
{\sc Wharington J.} (1998).
\newblock {\em Autonmous Control of Soaring Aircraft by Reinforcement
  Learning}.
\newblock PhD thesis, Royal Melbourne Institute of Technology.

\end{thebibliography}

\end{document}